# Just-Noticeable-Difference Based Edge Map Quality Measure

Ijaz Ahmad[1], and Seokjoo Shin[*]

Department of Computer Engineering,

Chosun University, Gwangju, 61452 South Korea

[1]ahmadijaz@chosun.kr, [*]sjshin@chosun.ac.kr (Corresponding author)

*Abstract*— The performance of an edge detector can be improved when assisted with an effective edge map quality measure. Several evaluation methods have been proposed resulting in different performance score for the same candidate edge map. However, an effective measure is the one that can be automated and which correlates with human judgement perceived quality of the edge map. Distance-based edge map measures are widely used for assessment of edge map quality. These methods consider distance and statistical properties of edge pixels to estimate a performance score. The existing methods can be automated; however, they lack perceptual features. This paper presents edge map quality measure based on Just-Noticeable-Difference (JND) feature of human visual system, to compensate the shortcomings of distance-based edge measures. For this purpose, we have designed constant stimulus experiment to measure the JND value for two spatial alternative. Experimental results show that JND based distance calculation outperforms existing distance-based measures according to subjective evaluation.

## I. Introduction

Edge detection characterizes boundaries of different objects in an image, which is a first step in analysis and understanding of image [1], [2]. It aids high-level computer vision tasks by substantially reducing the amount of information to be processed [3]. The reduction is done by extracting information from an input in such a way that the resultant image contains much fewer information than the original image [4], [5]. In addition, the extracted information is much more relevant to the other modules of an automatic vision system [4]. Since 60's, a variety of edge detection algorithms have been proposed [6], therefore it is necessary to evaluate the performance of these algorithms before being used in a computer vision system [1]. The common standard for a reliable edge map is that it contains all the relevant boundaries of an image as closely as possible, the detected edge pixels represent actual edges in the image and have certain degree of continuity without any disappearance of desired contours [1], [2]. Edge map measure estimates a performance score based on this standard. An effective measure is the one that can be automated and it's result correlates with human judgement perceived quality of the edge map [1], [4]. Edge detection evaluation algorithms can be used to improve performance of an edge detector by selecting suitable parameters for that algorithm, e.g. estimation of threshold value in last stage of edge detection [2], [4], [5].

Edge map evaluation is usually categorized into subjective or objective evaluation [7]. Subjective evaluation is the quantitative comparisons made by experts in that area using subjective ratings [2]. This approach does not give any statement about errors at pixel level. In addition, this method cannot be automated [3]. However, in most image processing applications human evaluation is the final step [7], e.g. performance of an edge map measure is evaluated using human mean opinion score [2]. On the other hand, objective evaluation is the qualitative comparison obtained from probabilistic measures, confusion metrics and figure of merits [5]. These methods calculate performance score at pixel level, thus information provided by these measures is important in automatic tuning of parameters for edge detection algorithms. Moreover, these methods can be automated [5].

Objective edge measures are further classified as non-reference objective measures and reference-based objective measures [2]. Non-reference measures are unsupervised evaluation and do not require a ground truth reference image [2], [3]. They only utilize the information from the obtained edge map and the original image itself to make the evaluation [2], [4]. Example of these methods are probabilistic measures, edge connectivity and width uniformity measures [3], [5]. The obvious advantage of these methods is that since they do not require any ground truth image, they can be used to evaluate the performance of edge detectors on non-synthetic images [5]. However, in their current stage, the use of non-reference edge measure is still restricted because edge quality evaluation is subjective and difficult to quantify without a ground truth [2]. The performance score of these methods does not clear how close the result is to desired contour [4]. The reference-based objective measures are supervised evaluation, which require a ground truth in addition to the resultant edge map to make their assessment [3], [5]. The comparison is made numerically using two techniques; assessment based on confusion metrics and assessment involves distance of misplaced pixels [1], [2], [8]. Confusion metric compares ground truth image and candidate edge map pixel by pixel, resulting in some ratio for presence or absence of edge pixels [2], [4], [8]. These methods ignore the issue of locality, thus the score suffers greatly when an edge is correctly identified but in a wrong location [2]. Distance-based assessment estimates a score by counting the number of erroneous pixels, and considers spatial distances of misplaced or undetected edge pixels as well [2], [6].

Distance-based edge evaluation methods compare two edge maps pixel by pixel, therefore, an edge pixel displaced even by a single pixel space is penalized. However, it is not the case in human visual system (HVS). HVS is only able to perceive a necessary magnitude of change between two stimulus to differentiate them [9]. This magnitude of difference is referred to as difference threshold, difference or differential limen, least perceptible difference, and just-noticeable-difference (JND) [9]. The term JND is widely adopted in psychology literature, and we will also use it to refer this phenomenon. In this work, JND based distance calculation is proposed to compensate the shortcomings of distance-based edge measures. We have performed experiments to estimate JND value for how small distance between two pixels can be perceived by a human. Then the JND value is used in classification of edges in the candidate image. The edges are classified as: correctly detected edges, misplaced edges with displacement less than JND value, misplaced edges with displacement greater than or equal to JND value and missed edge pixels. The performance score of existing distance-based measures falls behind the human judgement perceived quality of the edge map. One reason is that existing methods penalize the misplaced edges with displacement less than JND value, thus effects the overall performance score. Since, the displacement of misplaced edges less than the JND value is not perceivable by humans, therefore penalizing these edge pixels is not necessary in evaluation. In our proposed method, the performance efficiency is achieved by treating these pixels as correctly detected edges, while edge pixels with displacement greater than JND are penalized accordingly. The experimental results show the efficiency of JND based measure over other existing edge map measures according to subjective evaluation.

In this paper, we proposed a new objective edge map measure based on human perceptual property of JND. We have used force-choice experiment to estimate the JND value for distance between two pixels. The performance score of misplaced pixels whose distance is less than JND value is estimated as a ratio of correctly detected edges. The improvement in similarity search algorithm in the proposed method is attributed to the depth-constrained search and free and non-free edge pixels. These two properties of the algorithm avoids far away comparison and one-to-many relationship with false edge pixels, respectively. The performance comparison of proposed JND based edge measure is carried out with several existing objective

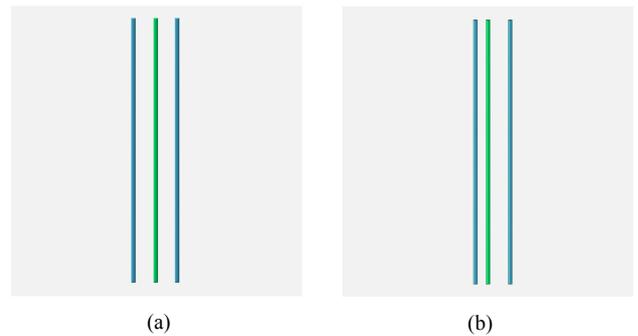

Fig. 1. Example of forced-choice trials in experiment to measure the JND value: (a) comparison stimulus on right with distance = 9 pixels (b) comparison stimulus on left with distance = 5 pixels.

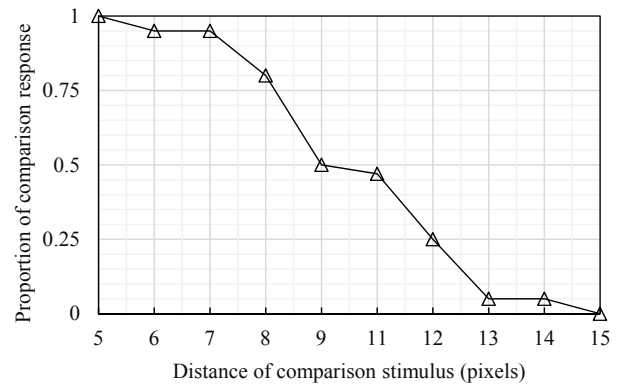

Fig. 2. Result of forced-choice experiment to simulate Just-Noticeable-Difference value for misplaced pixels.

measures by showing their correlation with mean opinion score (MOS) of edge maps.

The rest of the paper is summarized as follow. First of all, we conducted experiments to calculate JND value for distance between two pixels. Section II provides review of the experiment setup and protocol. Section III introduces edge map partitioning based on JND and search technique of the proposed method. Some experimental results on the synthetic data are given in section IV. Section V concludes the paper.

## II. JND Value Calculation

We have designed forced-choice experiment [10] to measure the JND value for two spatial alternative. The experiment consists of a line (constant stimulus), which is drawn at a constant distance of 10 pixels from a reference line. Another line (comparison stimulus) is drawn at a variable distance from the reference line in each trial. Fig. 1 shows the example of two trials of the experiment. Fig. 1 (a), the comparison stimulus is on the right of reference line with a distance equal to 9 pixels. Fig. 1 (b), the comparison stimulus is on the left side with a distance of 5 pixels from the reference line. The distance of comparison stimulus will vary from 5 pixels to 15 pixels, excluding distance equal to 10 pixels. The exclusion of 10 pixels distance guarantees no tie, and makes the subject to choose one stimulus closer to reference line every time. In each trial, both of the lines will appear alternatively on either left side or right side of the reference line. There are total of 20 possibilities for comparison stimulus and 10 trials for each possibility, hence total of 200 trials in the experiment. The primary focus of the experiment was to determine up to how many pixels a subject could perceive the difference in distance. To analyze the data we found the proportion of time that the comparison stimulus was selected closer to the reference line. The graph in Fig. 2 shows the proportion of times as a function of the comparison stimulus distance. The accuracy at distance 5 pixels and 15 pixels shows the reliability of the experiment. Both distance differences are large enough from 10 pixels that no error is made. On contrary, closer the distance of comparison stimulus getting to 10 pixels the probability of error increases i.e. for distance 9 pixels and 11 pixels the subject made many mistakes nearly 50% of the times. The JND value for distance between two pixels can be calculated as

$$\text{JND} = \frac{(L - M)}{2}, \quad (1)$$

where $L$ and $M$ are distances of comparison stimulus from reference line selected 25% and 75% of the times, respectively. The experimental results shows that the JND value for distance of pixels is 2 pixels. In other words, human perceived quality of edge map is above 2 pixels difference.

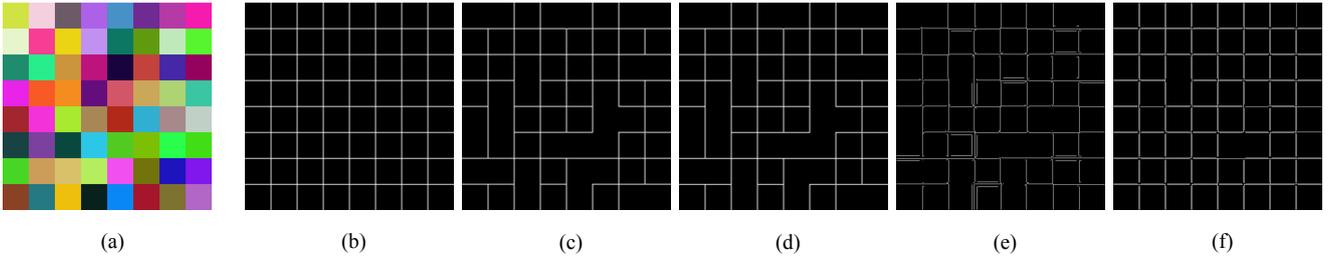

Fig. 3. (a) Original image. (b) Ground truth image. (c) Sobel edge map. (d) Prewitt edge map. (e) Laplacian of Gaussian edge map. (f) Canny edge map.

### III. PROPOSED METHOD

In our proposed method, based on the JND value calculated in section II, the candidate binary edge map is partitioned as: common edge pixels of ground truth and candidate edge map (i.e. correctly detected edges), edges with displacement from their ideal location (i.e. misplaced edges) and undetected edge pixels (i.e. holes in the candidate image). In this work, based on the JND value misplaced edges with displacement less than JND value are marked as correctly detected edges. This feature of algorithm makes our evaluation method based on important HVS property. In addition, the search for nearest edge pixel in candidate image is depth-constrained rather than sequential as shown in Fig. 4. For an edge pixel in Fig. 4 (a) only a range of pixels in Fig. 4 (b) is being searched. The motivation for depth-constrained search is to avoid matching of faraway false edge pixels. The proposed JND based edge measure for a ground truth ($G_t$) and detected edge map ($D_c$) is given as

$$\text{JNDBEM} = \frac{1}{\max(|G_t|, |D_c|)} \sum_{p=1}^{G_t} \frac{1}{1 + \alpha d^2 G_t(p)}, \quad (2)$$

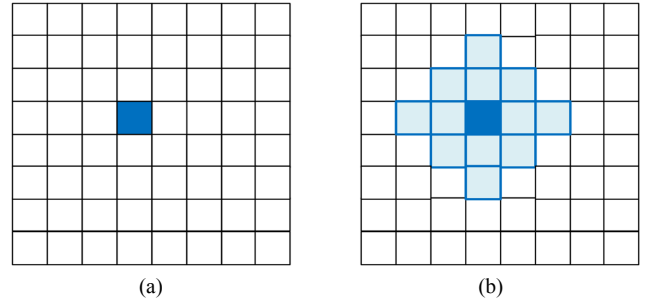

Fig. 4. Depth-constrained search for nearest edge pixels (a) Ground truth image. (b) Detected Edge map.

In Eq. (x) '|.|' represents cardinality of an image, '$p$' is the edge pixel of ground truth, '$\alpha$' is penalty parameter to tackle the false edge pixels of $D_c$ and '$d$' is a minimum distance between ideal edge pixel and an edge pixel in detected edge map. The value of '$\alpha$' in our experiments is set to 1/9 as suggested in [2], [6]. To avoid the contribution of one pixel multiple times to final score the matched elements are eliminated at every step. E.g., the correctly detected edge pixels are eliminated before the searching phase. In order to avoid many-to-one relationship of ground truth and detected edge pixels in case of under segmentation, the matched pixels of candidate image in the current search are eliminated for future search. For over-segmentation, this is achieved by the direction of search, i.e. for every undetected correct edge pixel of ground truth a misplaced edge pixel with minimum distance in candidate image is searched. The measure achieves a single quantitative index of an edge detector performance. For common edge pixels and misplaced edge pixels with displacement less than JND, the estimated performance score is the ratio of correctly detected edge pixels and maximum number of pixels in either of edge maps. On the other hand, for misplaced edge pixels with displacement greater than JND, the estimated score is based on the distance between ground truth and candidate edge map pixels. The score ranges

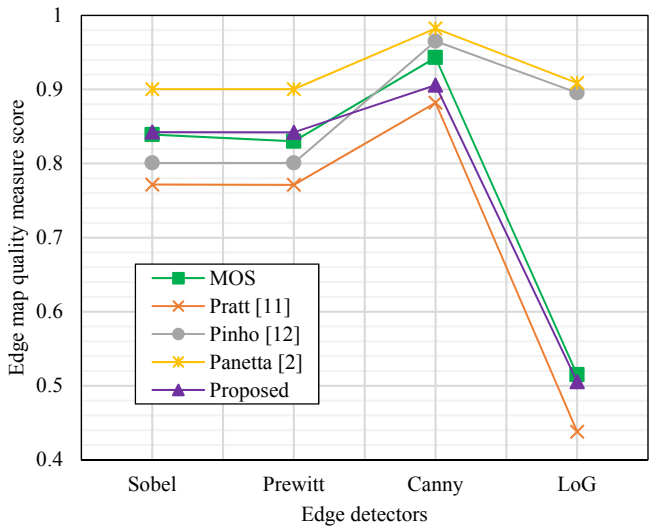

Fig. 5. Performance comparison of different edge map quality measures for various edge detection methods. The proposed measure score is highly correlated with the mean opinion score (MOS). LoG stands for Laplacian of Gaussian.

between 0 and 1, where 1 means a perfect match between candidate and ground truth edge map.

## IV. EXPERIMENTAL RESULTS

To evaluate performance of the proposed JND based edge map measure, we have conducted our experiments on synthetic data. The evaluation score of new measure is compared with three existing edge map evaluators; Pratt's FOM [11], Pinho's FOM [12] and Panetta's RBEM [2] edge measures. First, the edge maps are obtained using Sobel, Prewitt, Laplacian of Gaussian (LoG) and Canny edge detectors. We have used the parameters for these methods suggested in [2]. Fig. 3 is the tested synthetic image and its resultant edge maps. The performance score for these maps is obtained using the mentioned evaluation methods. A value 1 of these measures represents a perfect match between ground truth and candidate edge image. These measures are compared with subjective evaluation (Mean Opinion Score). The mean opinion score (MOS) is the average value of the ratings given to each edge map by human subjects. The ratings given are in range of 1 to 10, where 10 means the best quality.

The comparison of objective methods and normalized MOS is plotted in Fig. 5 for different edge detection algorithms. From the results, it is noteworthy that JND based edge measure score highly correlates with the subjective evaluation given by MOS for Sobel, Prewitt and LoG. However, for Canny edge detector score of the proposed method falls behind the Pinho's measure. The reason is that the Canny edge map is under-segmented, and since Pinho's measure does not consider the depth of search, therefore one-to-many relation for pixel comparison is made.

## V. CONCLUSION

In this paper, a new reference-based objective edge map measure has been proposed. The new measure incorporates just-noticeable-difference property of human visual system. For this purpose, experiments are being conducted to estimate the JND value for displacement of two pixels. To evaluate the proposed method, the performances of edge detection algorithms for synthetic data have been compared with MOS ranking. The results show that JND based edge measure outperforms existing methods with respect to subjective evaluation. The performance score of new measure highly correlates with the human perceived quality of edge map.

## ACKNOWLEDGMENT

This research was supported by the MIST(MInistry of Science & ICT), Korea, under the National Program for Excellence in SW supervised by the IITP(Institute for Information & communications Technology Promotion), (2017-0-00137).

## REFERENCES

[1] M. Baptiste, H. Abdulrahman, and P. Montesinos, "A Review of Supervised Edge Detection Evaluation Methods and an Objective Comparison of Filtering Gradient Computations Using Hysteresis Thresholds," J. Imaging, vol. 4, p. 74, May 2018, doi: 10.3390/jimaging4060074.
[2] K. Panetta, C. Gao, S. Agaian, and S. Nercessian, "A New Reference-Based Edge Map Quality Measure," IEEE Trans. Syst. Man Cybern. Syst., vol. 46, no. 11, pp. 1505–1517, Nov. 2016, doi: 10.1109/TSMC.2015.2503386.
[3] S. C. Nercessian, S. S. Agaian, and K. A. Panetta, "A new reference-based measure for objective edge map evaluation," Orlando, Florida, USA, May 2009, p. 73510J. doi: 10.1117/12.816519.
[4] H. Abdulrahman, B. Magnier, and P. Montesinos, "A New Normalized Supervised Edge Detection Evaluation," in Pattern Recognition and Image Analysis, vol. 10255, L. A. Alexandre, J. Salvador Sánchez, and J. M. F. Rodrigues, Eds. Cham: Springer International Publishing, 2017, pp. 203–213. doi: 10.1007/978-3-319-58838-4_23.
[5] S. Nercessian, K. Panetta, and S. Agaian, "A non-reference measure for objective edge map evaluation," in 2009 IEEE International Conference on Systems, Man and Cybernetics, San Antonio, TX, USA, Oct. 2009, pp. 4563–4568. doi: 10.1109/ICSMC.2009.5346779.
[6] B. Magnier, "Edge detection: a review of dissimilarity evaluations and a proposed normalized measure," Multimed. Tools Appl., vol. 77, no. 8, pp. 9489–9533, Apr. 2018, doi: 10.1007/s11042-017-5127-6.
[7] K. N. Plataniotis, "Comprehensive analysis of edge detection in color image processing," Opt. Eng., vol. 38, no. 4, p. 612, Apr. 1999, doi: 10.1117/1.602105.
[8] D. L. Wilson, A. Baddeley, and R. A. Owens, "A New Metric for Grey-Scale Image Comparison," Int. J. Comput. Vis., vol. 24, pp. 24–1, 1995.
[9] I. B. Weiner and W. E. Craighead, Eds., The Corsini Encyclopedia of Psychology. Hoboken, NJ, USA: John Wiley & Sons, Inc., 2010. doi: 10.1002/9780470479216.
[10] "Measuring Just-noticeable Differences Using Shape Size Changes | Sensation and Perception | JoVE." https://www.jove.com/v/10229/just-noticeable-differences (accessed Apr. 06, 2022).
[11] W. K. Pratt, Digital image processing, Third., vol. 242. New York, NY, USA: Wiley, 2001.
[12] A. J. Pinho and L. B. Almeida, "Edge detection filters based on artificial neural networks," in Image Analysis and Processing, 1995, pp. 159–164.